\documentclass[twoside,11pt]{article}

\usepackage{blindtext}
\usepackage{amsmath, amssymb}
\usepackage{enumitem}
\usepackage[T1]{fontenc}
\usepackage{lmodern}
\newlist{steps}{enumerate}{1}
\setlist[steps, 1]{label = Step \arabic*:}
\usepackage[dvipsnames]{xcolor}
\usepackage{float} 
\usepackage{caption} 
\usepackage{amsmath} 
\usepackage[abbrvbib]{jmlr2e}

\usepackage{lastpage}

\ShortHeadings{}{One and Two}
\firstpageno{1}

\begin{document}

\title{Adaptive KDE for Real-Time Thresholding: Prioritized Queues for Financial Crime Investigation }

\author{\name Danny Butvinik  \email butvinik@niceactimize.com \\
       \addr Chief Data Scientist\\
       NICE Actimize\\
       \AND
       \name Nana Boateng \email nboateng@ggc.edu \\
       \addr Senior Data Scientist\\
       NICE Actimize\\
       \AND
       \name Achi Hackmon \email achi.hackmon@niceactimize.com \\
       \addr Vice President, Head of AI \& Analytics\\
       NICE Actimize\\
       }

\maketitle

\begin{abstract}
 
We study the problem of converting a continuous stream of risk scores into stable decision thresholds under non-stationary score distributions. This problem arises in a wide range of detection systems where scores must be partitioned into prioritized processing regions while preserving semantic consistency over time.

We propose a density-anchored thresholding framework based on online adaptive kernel density estimation. The method summarizes the evolving score population in real time, identifies persistent local minima of the estimated density, and restricts threshold placement to these population-intrinsic separation points. Operational constraints are enforced by selecting thresholds whose induced tail mass satisfies prescribed limits, while anchoring decisions to stable structural features of the score distribution. 

The resulting thresholds are label-free, explainable, and robust to distributional drift, discretization effects, and transient fluctuations. The framework naturally extends to multiple thresholds, enabling the construction of ordered processing regions without altering score monotonicity.

Although motivated by financial-crime detection systems, the formulation is generic and applies to any scored stream requiring stable, capacity-constrained routing under evolving score populations. 
 
\end{abstract}

\begin{keywords}
  Kernel density estimation, anti-money laundering, unsupervised machine learning
\end{keywords}

\section{Introduction }
Modern detection systems emit a continuous stream of risk scores. Operations must convert this stream into one or more review queues under explicit intake constraints, while the score distribution itself is non-stationary, multimodal, and often discretized. In practice, this conversion is typically handled by a top-$K$ rule or by hand-tuned cutoffs. Such rules are fragile: small distributional drift can move a fixed cutoff from a sparse region into a dense one, creating abrupt swings in alert volume; adding secondary queues forces ad-hoc adjustments; and repeated manual retuning couples’ analysts’ workload to incidental fluctuations rather than to the underlying structure of the score space. Importantly, the difficulty is not merely operational but structural. The core challenge is to define decision boundaries that remain semantically stable as the score distribution evolves, while respecting externally imposed intake constraints. In this view, capacity does not determine where risk begins or ends; it constrains how much of an already-defined score region can be acted upon at any point in time. 

 We propose a density-anchored alternative. The score stream is summarized online by an adaptive kernel density estimator with boundary reflection, Epanechnikov kernels, and a pilot-guided local bandwidth (Sheather–Jones plug-in with Abramson’s transform). Thresholds are not set directly on scores but on the estimated density: we convert the density to a tail-mass curve to satisfy an intake target and then “snap” the cut to the nearest persistent valley (local minimum) detected across a range of bandwidths. This anchors decisions at natural separations of the population and yields stable intake even as the distribution drifts. The same mechanism produces multiple queues by solving two or more capacity targets and placing each boundary on a persistent valley.

The framework is streaming and label-free. Sliding windows or exponential forgetting provide controlled responsiveness; valley persistence acts as a regularizer that prevents over-reaction to transient bumps; optional hysteresis dampens cut motion further. Evaluation focuses on operational criteria—capacity adherence, intake variability, cut jitter, backlog behavior, and runtime—rather than on false-positive rates. Although motivated by financial-crime operations, the method is generic to any scored detector that must allocate scarce investigative capacity in real time.

This paper contributes: a formalization of density-anchored, capacity-matched thresholding for single and multi-queue routing; an online adaptive KDE with boundary corrections suitable for streaming; a persistence-based valley detector that stabilizes cuts without using labels; and an evaluation protocol on synthetic, drifting, multimodal streams demonstrating competitive capacity tracking with reduced threshold jitter and controlled backlog at $O(G)$ per-event update cost.

\section{ Mathematical Background and Related Work }

We model each transaction as a score $s \in [0,1]$ and the population by a density $f$ with distribution function $F$. KDE provides a nonparametric estimator $f$ from scores $s_{1} \cdots s_{n}$:
\begin{equation}
    \hat{f}(x) = \frac{1}{nh} \sum_{i=1}^n K\left(\frac{x-s_i}{h}\right)
\end{equation}
with bandwidth $h > 0$ and kernel $K$. On bounded support $[0,1]$ we use the Epanechnikov kernel, $K(u) = \frac{3}{4}(1-u^2)\mathbb{1}_{\{|u| \le 1\}}$,  because it minimizes the asymptotic mean integrated squared error (AMISE) among second-order kernels. Bandwidth governs the bias–variance tradeoff; the Sheather–Jones plug-in selects $h$ by minimizing an AMISE surrogate, replacing unknown functionals of $f$ by consistent pilot estimates. Bounded domains introduce edge bias; we correct it either by reflection (Include mirror points at $-s_i$ and $2-s_i$ in the sum when evaluating near $\{0, 1\}$ or by local-linear KDE, which removes $O(h)$ boundary bias while retaining $O(h^2)$ interior bias).

\begin{figure}[H] 
    \centering
    \includegraphics[width=0.8\textwidth]{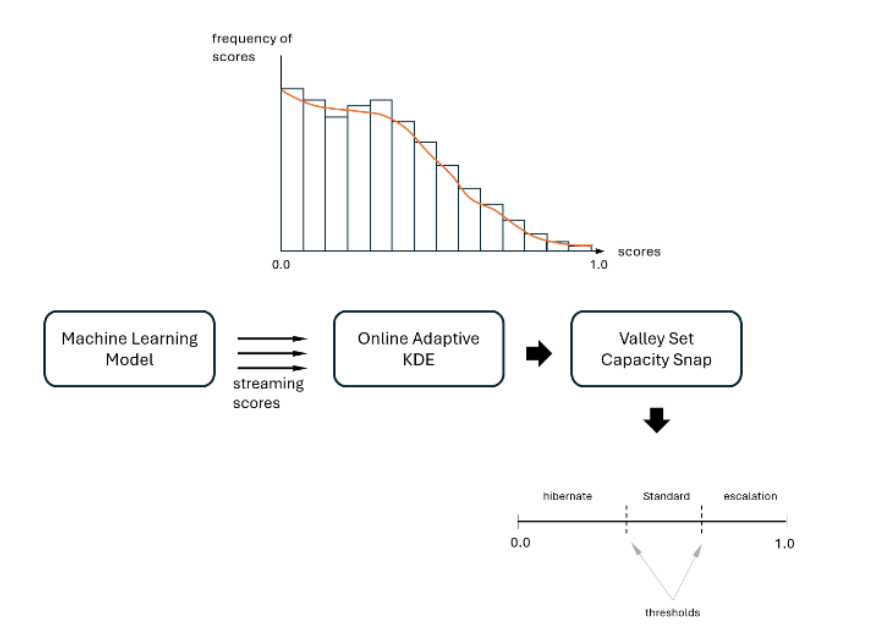}
    \caption{Visual Depiction of Method}
    \label{fig1_2026:fig2026}
\end{figure}

To reveal structure of varying granularity, we use adaptive KDE. With a positive pilot $\hat{f}$, Abramson’s square-root law sets a location-dependent bandwidth 

\begin{equation}
    h(x) = h_{0}\sqrt{\frac{g}{\hat{f(x)}} }
\end{equation}

where $h_{0}$ is a global scale and $g$ is the  geometric mean of $\hat{f}$ Bandwidth shrinks in dense regions and expands in sparse regions, sharpening genuine modes and valleys while damping spurious bumps. Derivatives of the density are estimated by differentiating the kernel: 

\begin{equation}
    \hat{f}^{m}(x) = \frac{1}{nh^{m+1}} \sum_{i}K^{m}\left( \frac{(x-s_i)}{h}\right)
\end{equation}

Local minima are points where $\hat{f}^{\prime}(x)=0$ and $\hat{f}^{\prime \prime} > 0$ these valleys act as natural cut candidates. As an alternative, one may partition 
$[0,1]$ via Bayesian Blocks: scores are binned along the axis and blocks are chosen to maximize a Poisson likelihood with a penalty for the number of blocks, yielding data-adapted change-points that typically coincide with density valleys.

\begin{figure}[H] 
    \centering
    \includegraphics[width=0.8\textwidth]{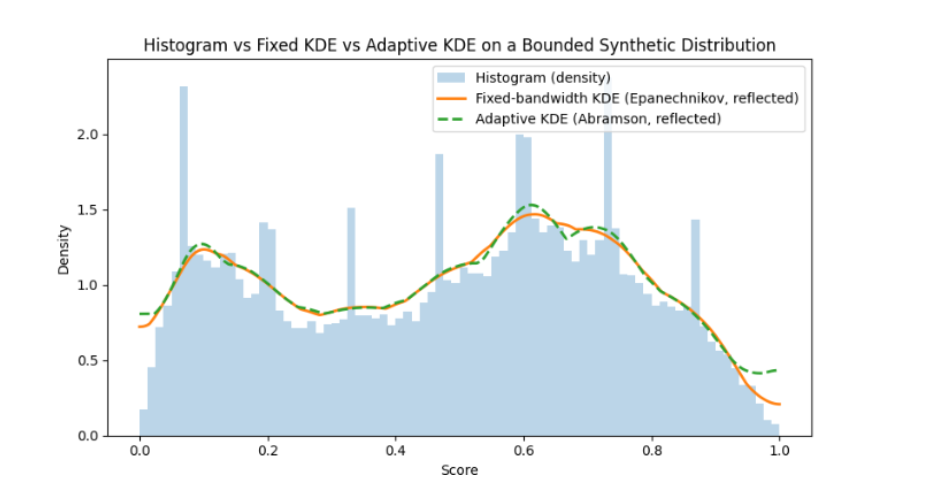}
    \caption{The illustration shows a single panel with histogram(density), fixed-bandwith Epanechnikov KDE with boundary reflection, and Adaptative of KDE(Abrahamson)}
    \label{fig2_2026:fig2026}
\end{figure}

Capacity enters through tail mass. For a target intake ratio 
$K \in (0,1)$ the capacity-true single threshold solves
$ \int_{t}^{1} f(x) dx =k $. Replacing $f$ by $\hat{f}$ yields the operational cut $\hat{t} = \hat{F^{-1}}(1-k)$. The sensitivity of alert volume to the cut is governed locally by $A'(t) = -n\,f(t)$. Placing cuts at density minima makes $f(t)$ small and therefore confers
stability: small movements in $t$ imply small changes in intake. For multiple queues with
cut set $c_1 \le \cdots \le c_m$, valleys provide the initial candidates; a final intra-interval
adjustment then meets the exact capacity.

For streaming scores $s_{t}$, we employ online KDE with either a sliding window of width $W$ or exponential forgetting with factor  On a fixed grid  $x \in [0,1]$

\begin{equation}
    \hat{f_{t}(x)} = (1-\alpha)\hat{f_{t-1}(x)} +\alpha K_{h_{t}}(s-x_{t})
\end{equation}

with boundary correction at each update. A pilot 
$\hat{f_{t}}$ updated likewise supplies 
$h_{t}(x) $ for the adaptive estimator. The grid makes per-event updates $O(G)$  with points and avoids storing the full history. Hysteresis and minimum-support constraints (update only when effective sample exceeds a threshold) prevent jitter in the cut sequence.

We draw the above on classical and modern results: Rosenblatt–Parzen KDE and AMISE theory; Epanechnikov optimality; boundary correction via reflection and local-linear estimators; plug-in bandwidth selection (Sheather–Jones); adaptive bandwidth selection (Abramson and successors); derivative KDE for mode/valley inference; and Bayesian Blocks for one-dimensional change-point partitioning. We specialize these tools to the unit interval, add online estimators for streams, and formalize stability via density-based sensitivity of capacity-true thresholds.

\section{Problem Formulation and Operating Assumptions}
    We consider a discrete-time stream of scored transactions for each business activity (BA) $b \in \mathcal{B}$. At time $t$, a BA produces scores $S_{b,t} = {s_{i} \in [0,1]}_{i=1} ^{N_{b,t}}$ drawn from an unknown distribution with CDF $F_{b,t}$ and density $f_{b,t}$. The bank specifies a target intake (capacity) $C_{b,t} \in \mathcal{N} $ per unit time and an operating mode with $m \in {1,2} $ thresholds (typically two or three queues: escalation / standard / hibernation).

\subsection{Decision variables and routing}
A routing policy for a BA at time $t$ is determined by a set of cut-points:

\begin{equation*}
     c_{b,t} = \{c_{1} < \cdots < c_{m}\} \subset  [0,1] 
\end{equation*}

with the understood convention that "escalation" receives scores in $[c_{m},1]$, "standard" (if present) receives  scores $[c_{m-1},c_{m})$, and all lower scores hibernate. Within any queue, ordering by score remains monotone (no re-ranking).


The induced expected queue sizes are
\[
A^{\uparrow}_{b,t}(\mathbf{c}) = N_{b,t} \int_{c_m}^{1} f_{b,t}(x)\,dx,
\qquad
A^{\circ}_{b,t}(\mathbf{c}) = \left\{
\begin{array}{ll}
N_{b,t} \int_{c_{m-1}}^{c_m} f_{b,t}(x)\,dx,  & \text{m=2} \\
0 & \text{m=1}
\end{array}
\right.
\]

Capacity adherence requires
\[
A^{\uparrow}_{b,t}(\mathbf{c}) + A^{\circ}_{b,t}(\mathbf{c})
\approx C_{b,t},
\]
either exactly or within a tolerance.

Thresholds define fixed score regions whose interpretation is invariant over time. Routing decisions operate only on the volume induced by these regions, not on their semantic meaning. This separation ensures that operational constraints do not implicitly redefine risk levels as workload or traffic fluctuates. 

\subsection{Stability and Candidate Sets}

Local volume sensitivity to a cut at $c$ is
\[
E_{b,t}(c) = N_{b,t} f_{b,t}(c).
\]
We seek stable operating points: small changes in $c$ should induce small changes in intake.
To that end, we restrict cut-points to a candidate set of valleys
\[
\mathcal{V}_{b,t}
= \left\{ x \in (0,1) : f'_{b,t}(x)=0,\; f''_{b,t}(x)>0 \right\},
\]
augmented, if needed, by one interior point within a selected interval to fine-tune capacity.
In practice, $f_{b,t}$ and $\mathcal{V}_{b,t}$ are estimated online
(Section~\ref{sec:online_kde}).

\subsection{Optimization Objective (Score-Only Regime)}

Given $C_{b,t}$, choose
\[
\mathbf{c}_{b,t} \in \mathcal{V}_{b,t}^m \cup \{\text{one interior fine-tune}\}
\]

\[
\underbrace{\min_{x \in X} \sum_{j=1}^{m} E_{b,t}(c_{j})}_{ \text{ stability(low elasticity) } } +
\underbrace{\lambda \left\| c-c_{b,t-1} \right\| }_{ \text{ hysteresis(low jitter) } } \text{s.t.} \bigl| A^{\uparrow}_{b,t}(\mathbf{c}) + A^{o}_{b,t}(\mathbf{c}) - C_{b,t} \bigr| \leq \delta
\]

with small $\lambda, \delta > 0$.
This yields capacity-true, valley-anchored thresholds with limited motion over time.

\subsection{Multi-Business Activities and Streaming Constraints}

\begin{itemize}
  \item \textbf{Per-BA independence.}
  Policies $\mathbf{c}_{b,t}$ are computed per BA from that BA’s score stream;
  distributional shapes may differ across BAs.

  \item \textbf{Streaming updates.}
  Thresholds update on a cadence (e.g., every $\Delta$ seconds) using an online estimator
  of $f_{b,t}$; a minimum effective sample size and hysteresis prevent jitter.

  \item \textbf{Guardrails.}
  Enforce minimum queue volumes, forbid cuts within $\varepsilon$ of the boundaries
  $\{0,1\}$, and fall back to a quantile cut when $\mathcal{V}_{b,t}$ is empty.
\end{itemize}

\subsection{Optional Label-Aware Refinement (Not Required)}

When band-level outcome estimates are available, define per-band utility
$U_{k,t}$ (e.g., expected true-positive yield per case) and replace the stability-only
objective with a bi-criterion selection that:
\begin{enumerate}
  \item meets capacity,
  \item keeps cuts at valleys, and
  \item maximizes $\sum_k U_{k,t}$
\end{enumerate}
subject to uncertainty bounds.
Absent labels, the score-only formulation above is adopted and no claim is made
about false-positive rates.

\section{Online Adaptive KDE on $[0,1]$}
\label{sec:online_kde}

Let a single business activity $b$ produce a score stream
$\{ s_{b,t} \in [0,1] \}_{t \ge 1}$.
At time $t$ we maintain a density estimate $f_{b,t}$ on a fixed grid
\[
X = \{x_j\}_{j=1}^G \subset [0,1].
\]
The kernel is Epanechnikov,
\[
K(u) = \frac{3}{4}(1-u^2)\mathbf{1}_{\{|u|\le 1\}}.
\]

Streaming updates use either a sliding window of the last $W$ scores or exponential
forgetting with factor $\alpha \in (0,1)$. With boundary reflection at both ends of $[0,1]$,
\[
\hat{f_{b,t}(x)}
= (1-\alpha) \hat{f_{b,t-1}}(x)
+ \alpha \left [ K_{h_{t}}(x)(x-s_{b,t} + K_{h_{t}}(x)(x+s_{b,t}) +K_{h_{t}}(x)(x-(2-s_{b,t}))  \right]
\]

evaluated at every grid point $x \in X$ The integral of $K_h$ is 1, and the reflected terms vanish in the interior so,
\[
\int_0^1 \hat{f_{b,t}}(x)\,dx = (1-\alpha) + \alpha =1
\]
at every step. A sliding-window estimator is recovered by replacing the convex update
with the average of the $W$ most recent reflected kernels.

Adaptive bandwidth follows Abramson’s square-root rule. A strictly positive pilot
density $\tilde f_{b,t}$ is updated the same way as $hat{f}_{b,t}$ but with a fixed global bandwidth $h_{0,t}$

Let  
\[
g_t = \exp\!\left( \int_0^1 \log \tilde f_{b,t}(x)\,dx \right)
\]
denote its geometric mean(computed on the grid with a small $\epsilon$ floor). Then
\[
h_t(x) = h_{0,t} \sqrt{ \frac{g_t}{\tilde f_{b,t}(x)} },
\]
shrinking in dense regions and expanding in sparse regions. We clip
$h_t(x) \in [h_{\min}, h_{\max}]$ to control variance and ensure numerical stability.

Boundary bias is handled by reflected by the online formula above, which is $\mathcal{O}(G)$ per event and requires
no local regression at the edges.  If desired, local-linear boundary correction can replace reflection on a subset of grid points; the streaming recursion is analogous, but reflection is typically preferred in real time because it is simpler and conserves mass exactly on $[0,1]$.

The global scale $h_{0,t}$ is re-estimated on a cadence (e.g., every few minutes). For a sliding window, $h_{0,t}$ is the Sheather–Jones plug-in computed from the current window; for exponential forgetting, set an effective sample size $n_{\mathrm{eff},t}  \approx \frac{1}{\alpha}$ (or its BA-specific variant) and use the standard AMISE plug-in with $n_{\mathrm{eff},t}$. In practice, a robust normal-reference initializer with variance capped by the Beta family on $[0,1]$ provides a stable starting value, with $S_J$ refinement when sufficient support accumulates.

Valley detection relies on the adaptive curve $\hat{f_{b,t}}$. On the grid, local minima are identified where the discrete first difference changes sign and the second difference is positive; this is equivalent to differentiating the kernel-based estimator in closed form, but finite differences are adequate at streaming cadence. These minima define candidate cutpoints for thresholding; their number and location evolve smoothly because both $\hat{f_{b,t}}$. and $h_{t}(.)$ evolve by convex updates with hysteresis coming from $\alpha$ and the cadence of $h_{0,t}$ refreshes.

\captionsetup{labelformat=empty}
\begin{figure}[H] 
    \centering

    \includegraphics[width=0.8\textwidth]{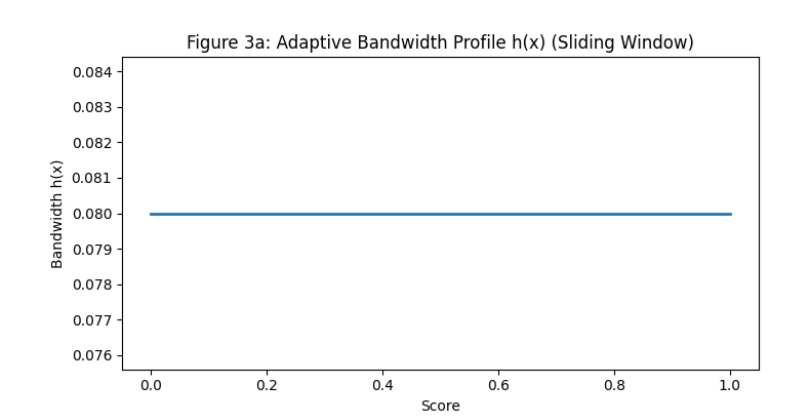}
\caption{Figure 3a: Bandwidth profile $h(x)$. The estimator tightens in dense regions and widens in
sparse tails, preserving real structure while dumping noise. Boundary reflection keeps the profile well-behaved near the end. }
    \label{fig3a_2026:fig2026}
\end{figure}
\captionsetup{labelformat=default}

\captionsetup{labelformat=empty}
\begin{figure}[H] 
    \centering

    \includegraphics[width=0.8\textwidth]{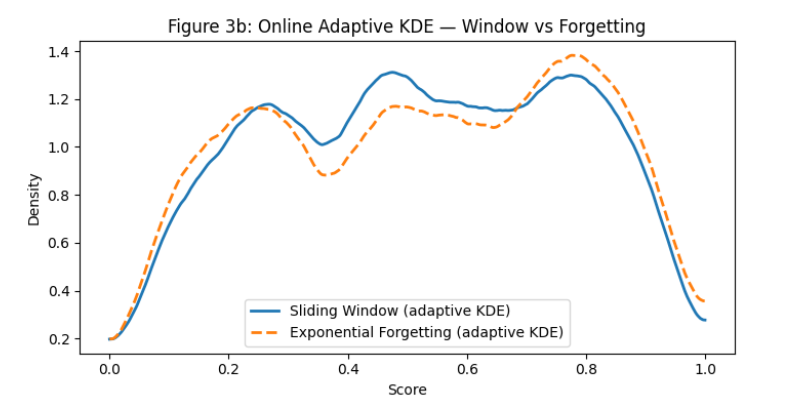}
\caption{Figure 3b: Sliding window vs exponential forgetting on the same stream segment; The window reacts faster to short shocks but is noisier, while forgetting yields a slightly smoother curve  with mid lag}
    \label{fig3b_2026:fig2026}
\end{figure}
\captionsetup{labelformat=default}

\captionsetup{labelformat=empty}
\begin{figure}[H] 
\centering

\includegraphics[width=0.8\textwidth]{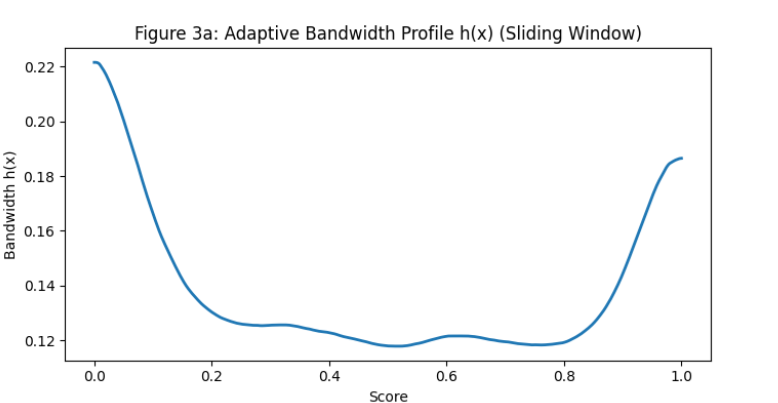}
\caption*{Figure 3c Capacity view buil from current estimate. The intake curve derived from the density(tail mass) is shown with the deployed threshold(s) overlaid. Cuts are snapped to the nearest valley and then fine-tuned within the boundary band to match}
    \label{fig3c_2026:fig2026}
\end{figure}
\captionsetup{labelformat=default}

Figure 3a, 3b, and 3c illustrate the online mechanics: adaptive bandwidths tailor smoothing to the local score regime (panel a), the streaming estimator can be run with either finite windows or exponential memory with predictable responsiveness/smoothness trade-offs (panel b), and thresholds are read directly from the current curve by snapping to valleys and nudging to capacity (panel c), enabling real-time, stable queueing.

Computationally, each arriving score updates the entire grid in $\mathcal{O}(G)$ time using
three reflected kernel stencils and the current $h_t(\cdot)$. Memory usage is
$\mathcal{O}(G)$ per BA for $f_{b,t}$, $\tilde f_{b,t}$, and $h_t(\cdot)$.
The method avoids binning, preserves normalization, respects the unit interval, and
produces an adaptive, smooth estimate whose valleys serve as stable anchors for capacity-true thresholds in real time.

\section{Valley Detection for Stable Thresholds}

Let $\hat{f_t}$ denote the adaptive, boundary-corrected kernel density estimate (KDE) on $[0,1]$ at time $t$.
A \emph{valley} is a local minimum of $\hat{f_t}$: points $v \in (0,1)$ where
\[
\hat{f}_t^{\prime}(v) = 0 \quad \text{and} \quad \hat {f_t}^{\prime \prime}(v) > 0 .
\]
With kernel derivatives

\[
\hat{f}_t^{t}(v) = \frac{1}{n_{t} h(x)^{m+1}} \sum_{i}K^{m}((x-s_{i})/h(x))
\]

under reflection, stationary points are obtained without finite-differencing error.
On a grid, we detect sign changes of $\hat{f_t}^{\prime}$ with a positivity check of $\hat{f_t}^{\prime \prime }$.

To suppress spurious minima induced by stochastic roughness, we require both \emph{scale} and \emph{salience}.
Scale is enforced by a SiZer-style consistency band: a candidate $v$ is retained only if the sign pattern
\[
\hat{f_t}^{\prime}\left(x \uparrow v) < 0, \qquad \hat{f_t}^{\prime}(x \downarrow v\right) > 0
\]
persists over a bandwidth neighborhood $h \in [h^-(v), h^+(v)]$.

Salience is measured by persistence against adjacent maxima.
If $u_L < u_R$ are the nearest maxima bracketing $v$, define
\[
S_t(v) = \min \{ \hat{f}_{t}(u_L) - \hat{f}_{t}(v), \; \hat{f}_{t}(u_R) - \hat{f}_{t}(v) \}.
\]
We retain $v$ when $S_{t}(v)$ exceeds a data-driven threshold obtained from a local variance proxy of $f_{t}$.
The surviving set $\mathcal{V}_t$ forms the admissible cut locations.

At the boundaries, reflection prevents artificial minima at $0$ and $1$; we additionally forbid
$\varepsilon$-neighborhoods of the endpoints.

\subsection*{Capacity-True Cuts}

Given a capacity ratio $\kappa_t$ (or count $C_t$), the capacity-true single cut solves
\[
\int_{t^\star}^{1} \hat{f_t}(x)\,dx = \kappa_t .
\]

For stability, we relocate to the nearest admissible valley $v \in \mathcal{V}_t$ satisfying
$v \le t^\star$ and minimizing local elasticity
\[
E_t(c) = N_t \hat{f}_t(c).
\]
If no such valley exists above a minimum-support threshold, we keep the interior cut at $t^\star$.

For two thresholds, we choose the highest pair $(v_1 < v_2) \subset \mathcal{V}_t$ maximizing stability,
subject to the mass of $[v_2,1]$ and $[v_1,v_2)$ meeting the escalation and escalation+standard targets.
A final intra-interval adjustment within $[v_1,v_2)$ matches the exact count.
This preserves score monotonicity inside each queue while anchoring boundaries at density minima.

\begin{figure}[H] 
    \centering

    \includegraphics[width=0.8\textwidth]{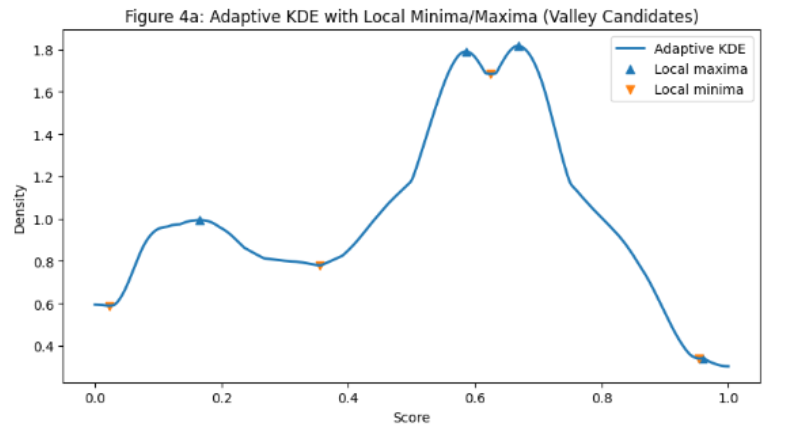}

    \label{fig4a_2026:fig2026}
\end{figure}

\begin{figure}[h]
\centering

\end{figure}

\captionsetup{labelformat=empty}
\begin{figure}[h]
\centering
\includegraphics[width=0.8\textwidth]{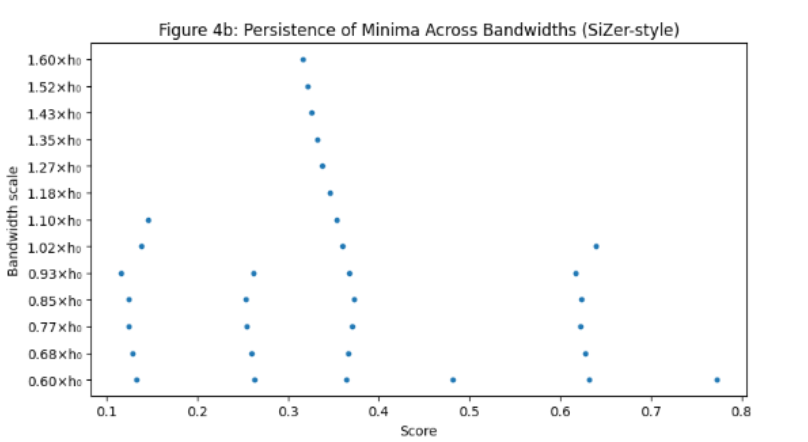}
\caption{Figure 4a and 4b:  Adaptive KDE with boundary reflection (left) identifies candidate thresholds at density valleys
(triangles). The persistence plot (right) shows which valleys survive across a range of bandwidths;
persistent valleys are preferred for stable, real-time operation.}
\end{figure}
\captionsetup{labelformat=default}

In streaming operation, $\mathcal{V}_t$ evolves smoothly because $\hat{f}_t$ updates by convex recursion.
We track valley identity by nearest-neighbor matching $v \mapsto v'$ across $(t-1) \rightarrow t$
with a maximum drift constraint and impose hysteresis:
a deployed cut moves only if the proposed replacement reduces elasticity by a fixed fraction
or the unconstrained capacity cut $t^\star$ crosses a midpoint between valleys.

Minimum-support guards drop valleys whose adjacent mass falls below a threshold, preventing cuts
supported by too few observations.
Under standard smoothness and bandwidth conditions, valley locations are consistent estimators
of the true minima, and the elasticity at deployed cuts converges to the population value,
yielding asymptotically stable capacity--intake behavior.

\section{Capacity Matching and Multi-Queue Routing}

Let $\hat{f}_t$ be the online, boundary-corrected adaptive KDE on $[0,1]$ for a given BA at time $t$,
and let $N_t$ be the volume over the operating window.
Define the tail mass

\[
U_{t}(c) = \int^{1}_{x} \hat{f_t}(u)du,\quad \hat{F}_{t}(x) = 1- U_{t}(x)
\]

\subsection{Single Threshold (One Queue)}

Given a target intake $C_t$ (count) or $\kappa_t = \frac{C_t}{N_t}$ (ratio), the capacity-true cut is the quantile

\[
t^\star = U_{t}^{-1}  ( \kappa_t ) = \hat{F_{t}^{-1}(1-\kappa_t)}
\]

For stability, we deploy
\[
c_t
= \arg\min_{v \in \mathcal{V}_t \cup \{t^\star\}}
\hat{f}_t(v)
\quad \text{s.t.} \quad
U_t(v) \ge \kappa_t
\]

i.e., snap to the nearest admissible valley with sufficient tail mass; if none exists, use $t^\star$.

When $U_t(c_t) > \kappa_t$ (overshoot due to snapping), trim inside $[c_t,1]$ by a within-band
percentile to hit $C_t$.
Local sensitivity of intake at the deployed cut is $\frac{\delta A_{t}}{\delta c}=- N_{t}\hat{f}_{t}(c)$, so valley anchoring directly minimizes elasticity.

\subsection{Two Thresholds (Three Queues)}

Let targets be $\kappa_t^{\uparrow}$ for Escalation and
$\kappa_t^{\uparrow+\circ}$ for Escalation + Standard, with
\[
\kappa_t^{\uparrow} \le \kappa_t^{\uparrow+\circ} \le 1 .
\]
Define the admissible valley set $\mathcal{V}_t$.
Choose 
\[
(c_t^{\circ}, c_t^{\uparrow})
=
\arg\min_{\substack{
v_1 < v_2  \in \mathcal{V}_t \cup \{t_1^\star, t_2^\star\}
}}
\left(
\hat{f}_t(v_1) + \hat{f}_t(v_2)
\right)
\]

subject to
\[
U_t(v_2) = \kappa_t^{\uparrow}, \qquad
U_t(v_1) = \kappa_t^{\uparrow+\circ}.
\]

If either constraint overshoots, apply an intra-interval fine-tune:
adjust only within $[v_1,v_2)$ (for Standard) or within $[v_2,1]$ (for Escalation)
by a small percentile shift to meet exact counts while keeping outer boundaries at valleys.
Routing is: 

$$
s > c_t^{\uparrow} \implies \textbf{Escalation}, \quad
c_t^{\circ} \le s \le c_t^{\uparrow} \implies \textbf{Standard}, \quad
s < c_t^{\circ} \implies \textbf{Hibernation}.
$$

with some  score-monotone ordering inside each queue.

\begin{figure}[h]
\centering

\includegraphics[width=0.8\textwidth]{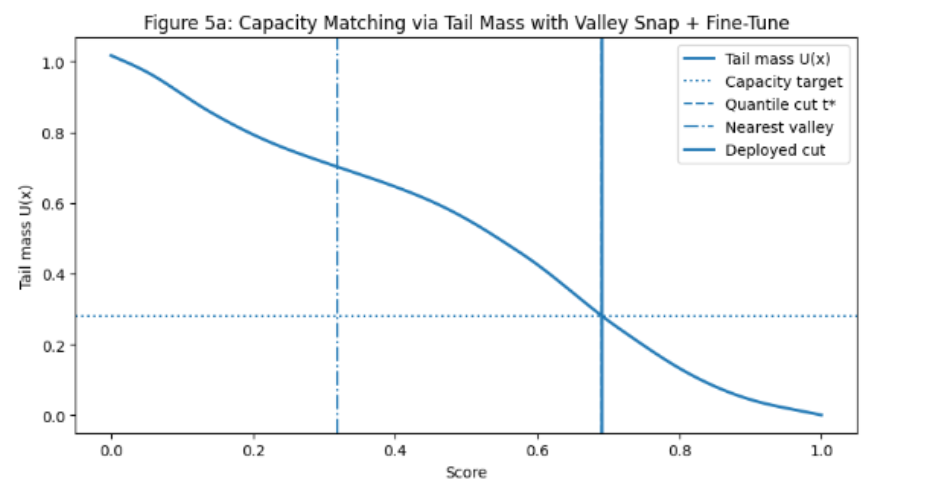}
\caption{}
\end{figure}

\captionsetup{labelformat=empty}
\begin{figure}[h]
\centering

\includegraphics[width=0.8\textwidth]{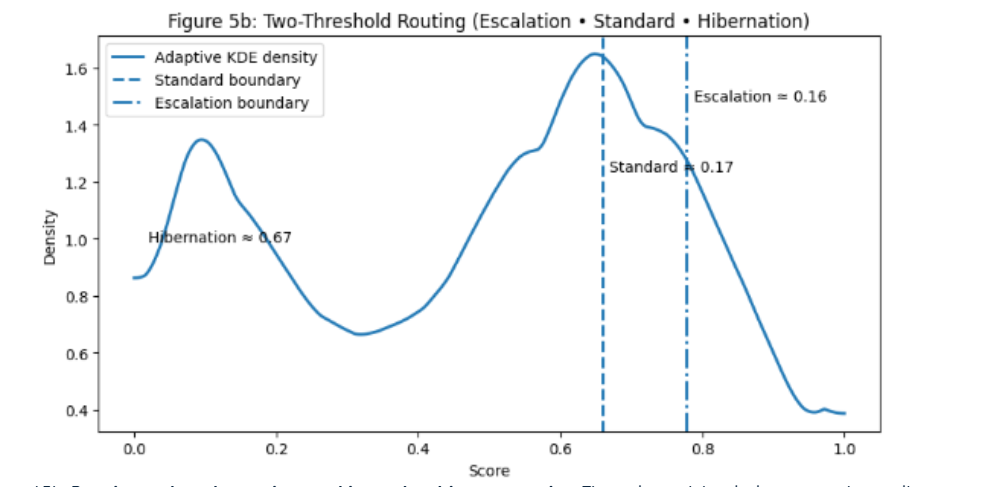}
\caption{Figure 5a and 5b: Density-anchored capacity matching and multi-queue routing.
The estimated density is converted to a tail-mass curve so cuts can be chosen directly at the stated capacity
and then snapped to nearby valleys for stability.With two cuts, the same principle partitions the population into Escalation,Satandard and Hibernation while preserving score-monotone ordering inside each queue}
\end{figure}
\captionsetup{labelformat=default}

Figure (5a) depicts a tail-mass curve translates the current density into “expected alerts above any cut.” We start at the capacity target, find the corresponding quantile cut, snap it to the nearest density valley to minimize volume sensitivity, and then make a small within-band adjustment so the expected intake exactly meets capacity. The deployed cut is therefore both capacity-true and stable.

Figure (5b) depicts two thresholds partition the density into Escalation, Standard, and Hibernation. Boundaries sit in valleys, so small models or traffic wiggles do not flip large numbers of cases across queues. The annotations show expected fractions per queue; within each queue, ordering remains purely by score, enabling staged intake and SLA-aware routing in real time.
The lowest-priority interval produced by this construction naturally serves as a deferred or hibernation region: cases are preserved for future consideration without altering score ordering or boundary semantics. This mechanism applies equally in detection settings where temporary load spikes must be absorbed without distorting score interpretation, including both financial-crime and fraud detection systems.

\subsection{Streaming Operation and Hysteresis}

Recompute $U_t$, $\mathcal{V}_t$, and proposed cuts on a fixed cadence.
Deploy updates only if they pass guardrails:
\begin{itemize}
\item \textbf{Hysteresis:} move a cut only if it reduces $\hat{f}_t(c)$ by at least a fraction $\eta$
or if the unconstrained quantile $t^\star$ crosses the midpoint to the next valley.
\item \textbf{Minimum support:} forbid cuts whose adjacent queue mass falls below a $\frac{M_\textbf{min}}{N_t}$
\item \textbf{Edge safety:} enforce $c \in [\varepsilon, 1-\varepsilon]$.
\item \textbf{Discretization:} if scores have ties or rounding, prefer valley locations that avoid
knife-edge boundaries; within-band trimming resolves remaining ties.
\end{itemize}

\subsection{Multi-BA Coordination (Optional)}

Apply the procedure per BA with its $\hat{f}_{b,t}$ and $C_{b,t}$.
If a global capacity $C_t$ must be split, allocate BA quotas by any external rule
(e.g., fixed proportions), then perform BA-local capacity matching.
Because cuts are valley-anchored within each BA, sensitivity and governance remain comparable.

\subsection{Outputs}

At each update, the system emits:
\begin{enumerate}
\item Deployed cut(s) $c_t$ (or $c_t^{\circ}, c_t^{\uparrow}$),
\item Expected counts per queue, $N_t U_t (c)$ and $N_t (U_t(c^\circ)-U_t (c \uparrow))$

\item Stability readouts $\hat{f}_t(c)$ at each cut,
\item An audit record of candidate valleys considered, fine-tune amount (if any), and and guardrails that bound the motion.
\end{enumerate}

\section{Experimental Evaluation on Synthetic Streams and Ablations}

We evaluate on label-agnostic, streaming scenarios that reflect capacity-driven routing.
The data generator produces per-BA score streams on $[0,1]$ as mixtures of Beta components
with occasional regime shifts, seasonal oscillations, and discretization to simulate rounding.
Three BAs are used with distinct shapes (unimodal skewed, bimodal with a narrow valley,
and trimodal with a crowded tail). Streams run for $T$ days with updates every $\Delta$ minutes.
Capacity targets $C_{b,t}$ vary slowly to mimic staffing changes and are also perturbed
by short bursts to test surge handling.
No class labels are needed for the primary metrics; a secondary, optional study attaches
synthetic outcomes to confirm that our policy preserves score monotonicity.

Baselines include batch top-$K$ (end-of-day), sliding-window quantiles (Greenwald--Khanna sketch), fixed-bandwidth KDE with reflection,
rolling EWMA cuts on raw counts, and non-reflected KDE on $[0,1]$ (to expose edge bias).
Our method uses an online adaptive KDE with reflection, valley snapping,
within-band fine-tuning to match capacity, and hysteresis.

\captionsetup{labelformat=empty}
\begin{figure}[H]
\centering
\includegraphics[width=0.8\textwidth]{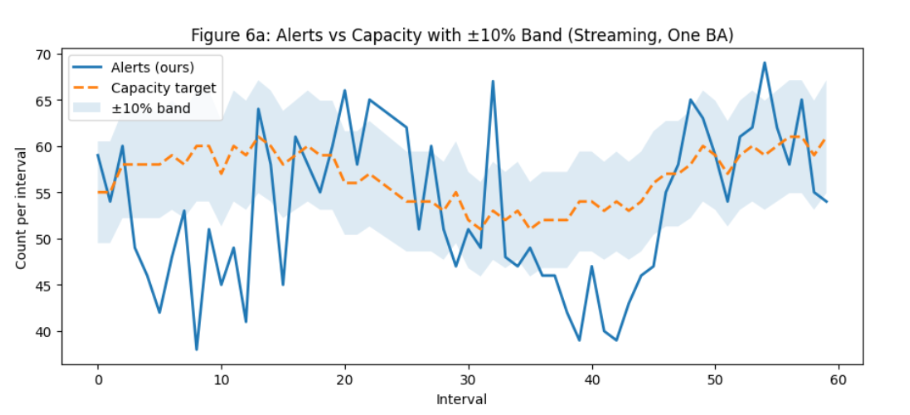}
\includegraphics[width=0.8\textwidth]{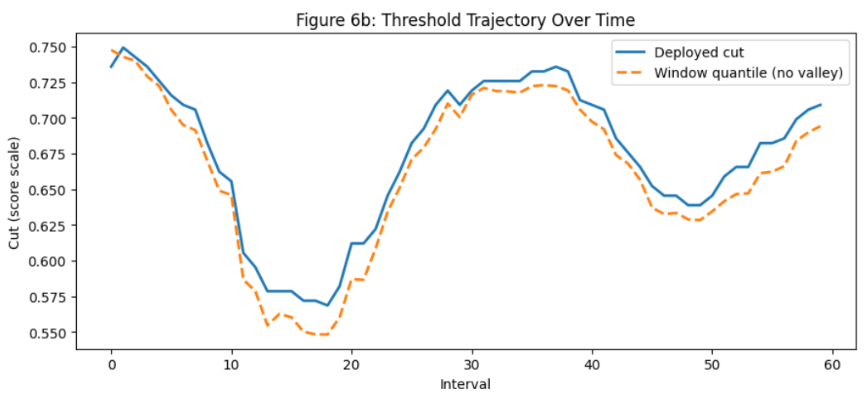}
\includegraphics[width=0.8\textwidth]{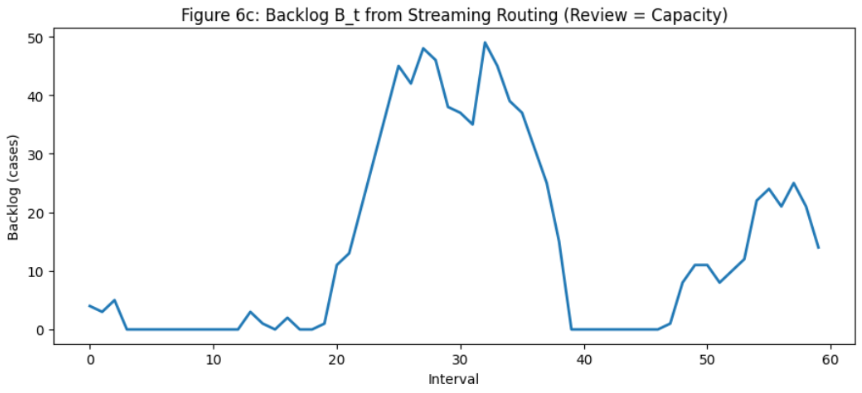}
\caption{Figure 6a, 6b and 6c: Compact view of the streaming policy; density-anchored thresholds keep intake on target and backlog controlled as scores drift -- showing volumes vs capacity ,threshold motion, resulting queue dynamics}
\end{figure}
\captionsetup{labelformat=default}

Figures~6a--6c illustrate how the method tracks a stated intake in real time.
Figure~6a shows realized alerts alongside capacity with a $\pm 10\%$ tolerance.
Figure~6b shows threshold motion, including a plain window-quantile cut for context.
Figure~6c simulates backlog, confirming that stable intake prevents accumulation under routine variability.

The central metric is capacity adherence.
For each BA and interval we record alerts $A_{b,t}$ and targets $C_{b,t}$ and summarize
$|A_{b,t}-C_{b,t}|$, relative deviation $|A_{b,t}-C_{b,t}|/C_{b,t}$,
and the fraction within a tolerance band.
Stability is measured by intake variability (coefficient of variation of $A_{b,t}$),
cut jitter $|c_{b,t}-c_{b,t-1}|$, and elasticity at deployment $\hat{f}_{b,t}(c_{b,t})$ as a local sensitivity proxy.

Queue health is assessed via  a simple backlog simulator
\[
B_t = \max\{0,\, B_{t-1} + A_t - R_t\},
\]
with fixed review capacity $R_t$.
We report the excedance probability $\Pr(B_t > \beta C)$ and mean time between breaches.

Cross-BA portability is summarized by the dispersion of elasticities across BAs and by the proportion of cuts anchored at valleys. For discretized scores we additionally report tie-break volatility (fraction of cases flipping in or out of the queue due to ties near the cut).

Ablations isolate the contribution of each design choice. We compare adaptive versus fixed bandwidth; reflection versus no boundary correction; valley snapping versus plain quantile cuts; and hysteresis on versus off. We also vary the update cadence, effective window size, and minimum-support thresholds to map accuracy–stability–latency trade-offs. Stress tests introduce tail explosions (short-lived high-score surges), valley vanishing (two modes merging), and rounding shifts (coarser discretization) to test whether cuts migrate to new stable locations without overshooting capacity.

An optional label-aware check uses synthetic outcomes (monotone in the true score with controlled miscalibration and cohort mixing) to confirm two invariants: ordering within each queue remains monotone in the score, and any operational stability gains do not depend on labels.

All experiments report medians and interquartile ranges over random seeds.
Runtime is profiled as wall-clock update time per event versus grid size $G$
and number of BAs, demonstrating $\mathcal{O}(G)$ per-event updates with constant
memory per BA.

\section{Limitations and Future Work}

This study is score-only. We deliberately avoid outcomes, calibration curves, and cost models; consequently, the method cannot claim improvements in precision or false-positive rates. Its guarantees are operational: capacity adherence and stability via valley-anchored cuts. Any re-ordering of risk would require label information or explicit utility models; we leave that outside scope.

Valley detection is data dependent. In sparse windows or heavily discretized score streams, apparent minima may be unreliable or vanish temporarily. Our guardrails (minimum support, hysteresis, edge exclusion) reduce jitter but also introduce latency; short shocks can be under-reacted to, while very small valleys can be ignored. Boundary correction by reflection is fast and mass-conserving on $[0,1]$, yet it can bias shapes if true densities accumulate at the edges; local-linear streaming analogues are possible but heavier.

Online estimation introduces design knobs. Window length or decay rate, cadence of bandwidth refresh, and the strength of hysteresis each trade responsiveness for stability. Although we show $O(G)$ per-event updates suffice in one-dimension, multi-BA deployments multiply memory and CPU, and extreme burst rates may require sketching or hardware acceleration. Score discretization and ties remain a practical nuisance around cuts; within-band trimming mitigates knife-edge effects but does not eliminate them when vendors emit coarse scores.

The formulation is per-cohort and per Business Activity (BA). Cross-BA capacity allocation is treated as exogenous; the paper does not solve joint optimization across activities or tenants. Governance is addressed operationally (audit trails for cut placement), not statistically: we do not yet attach uncertainty to deployed thresholds or to valley locations, which limit formal change-control arguments.
Several directions follow naturally. First, label-aware but conservative refinements:
per-band yield estimation with credible intervals, positive--unlabeled corrections for
review bias, and density-ratio policies that introduce narrow \emph{micro-segments}
only where repeated evidence accumulates without disturbing global monotonicity.
Second, principled uncertainty for structure: SiZer-style persistence tests or bootstrap
confidence bands for valleys in the streaming setting, together with stability theorems
that bound intake sensitivity under windowed KDE with decay.
Third, beyond one dimension: conditional densities $f(s \mid z)$, where $z$ captures
business activity or operational covariates; copula or projection-pursuit variants that
segment on a small number of auxiliary features while retaining a one-dimensional cut
policy.
Fourth, systems work: sketches for fast online KDE, automatic guardrail tuning, and the
addition of differential-privacy noise for regulated environments.
Fifth, policy coupling: integration of capacity matching with queueing control (backlog
dynamics and service-level constraints), and the study of multi-threshold policies as
stochastic controllers with performance bounds.
Finally, benchmarking: a public synthetic suite for scored-stream queueing with agreed
baselines (top-$K$, streaming quantiles, fixed-bandwidth KDE) and standardized stability
metrics to enable reproducible comparisons.

\section{Conclusion}

We presented a density-anchored, capacity-aware method for building prioritized queues from streaming risk scores. The approach treats the score stream as a one-dimensional stochastic object, maintains an online adaptive KDE with boundary correction, and uses the resulting valleys as stable, explainable candidates for thresholds. Capacity is met by snapping cuts to these natural boundaries and, only if needed, making a small intra-band adjustment. The policy preserves monotone ordering by score within each queue, supports single or multiple thresholds for escalation/standard/hibernation, and operates in real time without manual trial-and-error.

Throughout, capacity is treated as a constraint applied to structurally defined score regions, rather than as a driver of threshold placement.

The value is operational and governance centric. Anchoring cuts at density minima reduces sensitivity of intake to small setting changes, yielding predictable volumes, smoother queues, and portable behavior across cohorts and business activities. Because the machinery is nonparametric and score-only, it travels across tenants and systems, producing an audit trail for why thresholds sit where they do. In comparative experiments on synthetic streams, the method matches capacity as reliably as quantile baselines while exhibiting markedly lower volatility and cut jitter, especially under discretization and transient surges.

We do not claim improvements in precision absent outcomes; instead, we provide a transparent mechanism for real-time routing that is robust to incidental fluctuations and easy to justify. Future extensions—band-level yield estimation with uncertainty, density-ratio novelty signals, conditional segmentation with a few operational covariates, and formal queueing-control coupling—can add selective label awareness and broader policy intelligence without abandoning the core principle: discover structure first, then route along it.

\end{document}